\documentclass[letterpaper,twocolumn]{article} 

\usepackage{ist}
\usepackage[fleqn]{amsmath}
\usepackage[utf8]{inputenc} % allow utf-8 input
\usepackage[T1]{fontenc}    % use 8-bit T1 fonts
\usepackage{url}            % simple URL typesetting
\usepackage{booktabs}       % professional-quality tables
\usepackage{mathrsfs}       % https://www.ctan.org/pkg/mathrsfs
\usepackage{amsfonts}       % blackboard math symbols
\usepackage{nicefrac}       % compact symbols for 1/2, etc.
\usepackage{microtype}      % microtypography
\usepackage{titlesec} 
\usepackage{array}
\usepackage{authblk}
\usepackage{placeins}
\usepackage{xspace}
\usepackage{setspace}

\usepackage{makecell}
\usepackage{array}
\usepackage{multirow}
\usepackage{enumitem}% http://ctan.org/pkg/enumitem
\usepackage{dblfloatfix}

\DeclareSymbolFont{symbols}{OMS}{cmsy}{m}{n}
\DeclareSymbolFontAlphabet{\mathcal}{symbols}

\usepackage{hyperref}
\makeatletter
\def\HyPsd@expand@utfvii{}
\makeatother
\usepackage{xcolor}
\hypersetup{
 pdfpagemode=UseNone,
 pageanchor=true,
 plainpages=false,       % for problems with page referencing
 hypertexnames=false,    % for handling subfigures correctly
 bookmarksnumbered=false,% include the section numbers in the list
 bookmarksopen=true,     % In the list, display highest level only
 bookmarksopenlevel=3,   % display three levels of bookmarks
 pdfpagemode=UseNone,    % show just the page
 pdfstartview=Fit,       % defail page view is the whole page at once
 pdfborder={0 0 0},      % we don't want those silly boxes for links
 breaklinks=true,        % allow breaking links
 colorlinks=false,       % don't change coloring of links
 nesting=true,
 linktocpage
}

\title{Adapting Pretrained Networks for Image Quality Assessment on High Dynamic Range Displays}

\author[1]{Andrei Chubarau}
\author[2]{Hyunjin Yoo}
\author[2]{Tara Akhavan}
\author[1]{James Clark}
\affil[1]{Department of Electrical and Computer Engineering, McGill University, Montreal, Canada}
\affil[2]{Faurecia IRYStec Inc., Montreal, Canada}
\affil[ ]{\textit{andrei.chubarau@mail.mcgill.ca, \{hyunjin.yoo, tara.akhavan\}@forvia.com, james.j.clark@mcgill.ca}}

\date{} % date has an empty field.

\begin{document} 

\maketitle{}

\newcommand{\candela}{cd/m$^2$\xspace}
\newcommand*{\subsubsubsection}[1]{\textbf{\vspace{-10pt}\newline\indent\em#1.\hspace{1pt}}}

\thispagestyle{empty} % prevents the first page to be numbered

%%%%%%%%%%%%%%%%%%%%%%%%%%%%%%%%%%
% Abstract
%%%%%%%%%%%%%%%%%%%%%%%%%%%%%%%%%%

\begin{abstract}
Conventional image quality metrics (IQMs), such as PSNR and SSIM, are designed for perceptually uniform gamma-encoded pixel values and cannot be directly applied to perceptually non-uniform linear high-dynamic-range (HDR) colors. Similarly, most of the available datasets consist of standard-dynamic-range (SDR) images collected in standard and possibly uncontrolled viewing conditions. Popular pre-trained neural networks are likewise intended for SDR inputs, restricting their direct application to HDR content. On the other hand, training HDR models from scratch is challenging due to limited available HDR data. In this work, we explore more effective approaches for training deep learning-based models for image quality assessment (IQA) on HDR data. We leverage networks pre-trained on SDR data (source domain) and re-target these models to HDR (target domain) with additional fine-tuning and domain adaptation. We validate our methods on the available HDR IQA datasets, demonstrating that models trained with our combined recipe outperform previous baselines, converge much quicker, and reliably generalize to HDR inputs. 
% We also conduct a small-scale subjective IQA experiment on HDR displays for more focused evaluations, confirming our findings.
% confirm that domain adaptation reduces the degradation in performance associated with the domain shift from SDR to HDR and the inclusion of non-standard viewing conditions, 
% domain adaptation significantly outperform baselines without domain adaptation
\end{abstract}

\section{Introduction}
\label{sec:introduction}

Real-world scenes are brighter and more vivid than their digital twin reproductions. While 8-bit gamma-encoded color values drive the common standard-dynamic-range (SDR) displays, high-dynamic-range (HDR) imaging enhances the viewing experience by encoding a significantly wider range of luminance with more precision, allowing to represent a larger portion of the visible color gamut. Despite its advantages, HDR also brings complexity to the imaging pipeline. The vast majority of applications and algorithms operate on SDR content and do not yet extend to HDR. 

Image quality is a critical performance metric in all visual applications, be they SDR or HDR. Classical image quality assessment (IQA) relies on hand-crafted mechanisms based on mathematical models of the human visual system (HVS). With deep learning \cite{deeplearning}, IQA has evolved toward jointly optimizing feature representations and inference directly from image data. Training deep image quality metrics (IQMs) from scratch, however, is a challenging task, because IQA datasets are limited in size \cite{kadis700k}, especially for HDR \cite{upiq}. Recent methods address overfitting and handle large images by dividing inputs into smaller patches, computing and aggregating patch-wise metrics. Initially, IQMs based on convolutional neural networks (CNNs) essentially processed patches independently \cite{WaDIQaM, pieapp}; the current state-of-the-art models take advantage of transformer architecture \cite{attentionIsAllYouNeed, iqt, musiq, vtamiq}, or combinations of CNNs and transformers \cite{ahiq}, to capture more complex global interdependencies between patch-wise inputs.

% While convolutional neural network (CNN)-based IQMs tend to process patches independently \cite{WaDIQaM, pieapp}, state-of-the-art transformer (\cite{attentionIsAllYouNeed, transformerSurvey})-based models \cite{iqt, musiq, vtamiq}, or combinations of CNNs and transformers \cite{ahiq}, capture more complex global interdependencies between patch-wise inputs.

Only a handful of traditional IQMs, and even fewer deep learning-based IQMs, are natively designed for HDR content. Most notably, HDR-VDP \cite{hdrvdp2_2011, hdrvdp3}, a metric that models contrast detection for a wide range of luminance conditions, has achieved wide recognition. Most SDR algorithms, on the other hand, rely on the perceptual uniformity of gamma-encoded sRGB images, making them unsuitable for accurately processing perceptually non-uniform HDR color values. While perceptually uniform (PU) encoding \cite{puEncoding2008, pu21} and perceptual quantizer (PQ) \cite{pq} transform linear photometric color values into perceptually uniform units, thereby enabling the application of some SDR metrics, the effectiveness of SDR methods on HDR images is limited. With the scarcity of HDR data, most deep learning methods similarly operate primarily on SDR content. Previous attempts at HDR IQA involve training networks from scratch on PU-encoded data \cite{upiq, biqaHDR}, overlooking the clear advantages of transfer learning \cite{transferLearning2011}.

In this paper, we investigate more effective strategies for training deep learning-based IQA models for HDR\footnote{\url{https://github.com/ch-andrei/HDR-IQA-dom-adapt}}. Instead of training on HDR data from scratch, we leverage networks pre-trained on SDR content and propose special fine-tuning strategies to re-target such networks to HDR. First, we explore several modifications to the training procedure with PU-encoded units to facilitate transfer learning. Second, we train with domain adaptation (DA) to reduce the degradation in performance associated with the domain shift from SDR to HDR. Third, while we focus on IQA for HDR, we aim to provide adequate performance for SDR and HDR data, which allows for flexible usage of the trained models in real-world applications of IQA. We validate our findings by retraining PieAPP \cite{pieapp} and VTAMIQ \cite{vtamiq} to outperform previous baselines in HDR IQA on the available datasets (SDR and HDR).
% as well as on our own small-scale subjective IQA experiment on HDR displays. 
Our experiments emphasize the importance of transfer learning, 
% which we facilitate with better normalization and domain adaptation, 
as demonstrated by stronger generalization on both SDR and HDR data.

% simulating display-referred values for SDR data where possible.

% The novel contributions are i) specialized use of domain adaptation to retarget networks pre-trained on SDR data to HDR applications (instead of training from scratch), ii) our improvements to the procedure for training deep networks with PU-encoded color values, iii) validation of our strategies on existing SDR/HDR IQA datasets and our own subjective data for IQA on HDR displays, iv) ablations on our models and training procedures. Our code and data will be publicly available.

\section{Related Work}

% A brief overview of relevant work is provided with a focus on HDR imaging and Full-Reference (FR) IQA. 
% HDR stuff:
% PU encoding details
% Display model simulation equations etc
% IQA stuff:
% Basics for FR NR IQA
% VTAMIQ, MUSIC
% HDR IQA previous attempts

\subsection{Image Quality Assessment}

Conventional full-reference (FR) IQA correlates image quality with the perceptual difference between a reference and a distorted image. The comparison can be based on error visibility \cite{nqm2000, vsnr2007}, structural similarity \cite{ssim2004, msssim2003, iwssim2011}, information content \cite{ifc2005, vif2006}, contrast visibility \cite{hdrvdp2005, hdrvdp2_2011}, or various other feature similarities inspired by the Human Visual System (HVS) \cite{fsim2011, GMSD2013, mdsi2016, haarPSI2016, dog-ssim} and optionally modulated by visual saliency \cite{vsi2014, deepgaze}. More recent work uses deep learning \cite{deeplearning} for data-driven IQA. Instead of hand-crafted features, deep FR IQMs typically compare deep layer activations for two images \cite{cnnsIqa2016, lpips2018, WaDIQaM, pieapp}. The training is done by optimizing mean absolute error (MAE) or mean squared error (MSE) between the predicted and the expected quality values, optionally using additional guidance by pairwise preference \cite{pieapp} or ranking loss \cite{RankIQA, lpips2018}. With limited data, the advantages of transfer learning \cite{transferLearning2011, transferLearning} motivate the use of feature extraction networks initially trained for other vision tasks (e.g., classification or segmentation \cite{imageNet}) and subsequently fine-tuned to IQA. 

Because CNN-based methods tend to restrict input image resolution, recent work uses patch-wise processing for more flexibility: an image is split into smaller patches (randomly sampled or tiled), and patch-wise quality scores are computed and combined with averaging or weighted pooling \cite{deepQa, WaDIQaM}. As a notable example, the PieAPP quality metric \cite{pieapp} predicts individual quality scores and the corresponding patch weights for patches of $64\times64$ pixels. The use of patches further allows for data augmentation: a sequence of randomly sampled patches offers a reasonably novel ``view'' of the same data. However, for real-world applications with HD images, this can quickly become computationally intensive as the number of patches increases to cover more pixels. Lastly, transformers instinctively conform to patch-wise IQA, because the transformer architecture \cite{attentionIsAllYouNeed, vit2020} natively uses sequences as inputs. Among recent work on transformer-based IQMs, MUSIQ \cite{musiq} and VTAMIQ \cite{vtamiq} employ multi-scale patch processing to adapt to large resolution inputs common in practical applications of IQA. 
% The vast majority of IQA metrics are intended for SDR content. 
% with the advent of larger IQA datasets, notably KADID-10k \cite{kadid10k} and PieAPP \cite{pieapp}, training deep IQMs from scratch has become more feasible.
% For instance, PieAPP quality metric \cite{pieapp} requires input images to be tiled into $64\times64$ patches, which can be very computationally intensive for HD images.
% However, because real-world applications use large resolution images, small-scale patch-wise IQMs are often suboptimal \cite{musiq}. CNN-based methods tend to further restrict input image resolution. Conversely, some transformer-based IQMs, such as MUSIQ \cite{musiq} and VTAMIQ \cite{vtamiq}, are natively adapted to multi-scale inputs, making them more effective for modelling real-world imagery.

\subsection{Representing Real-World Displays}
\label{sec:displayModel}

% For displayed-encoded HDR content, the Perceptual Quantizer (PQ) \cite{pq}
Although the viewing experience varies widely according to viewing conditions and across different displays \cite{lrt2014, puEncoding2008}, many computer vision algorithms operate directly on 8-bit gamma-encoded sRGB color values designed for cathode-ray tube (CRT) displays with around 100 \candela peak luminance. HDR displays, on the other hand, depict a significantly wider range of visible color with luminance levels that reach 5000 \candela. To describe both SDR and HDR content on a unified scale, it is convenient to represent visual content in physical units of luminance emitted by a display as modeled by the gain-offset-gamma model \cite{BERNS1996}:
\begin{equation}
\label{eqn:displayModel}
    L = (L_{max} - L_{blk}) F(V) + L_{blk},
\end{equation}
where $L$ is the emitted luminance in \candela, $L_{max}$ and $L_{blk}$ are the maximum and the black level luminance of the display in \candela, $V$ is the display-encoded luma in the range 0--1, and $F$ is the EOTF, the inverse of the opto-electronic transfer function (OETF). For SDR, $F(V)=V^\gamma$, where $\gamma$ is the gamma-correction parameter (typically, $\gamma=2.2$), or the sRGB non-linearity. For HDR, $F$ can be Hybrid Log Gamma \cite{hlg}, PQ \cite{pq}, or $F(V)$ can directly encode linear scene luminance. 

We can optionally extend \autoref{eqn:displayModel} to account for ambient light reflected from the display \cite{hdriMantiuk} by adding the ambient reflection term $L_{amb}$, computed as
\begin{equation}
\label{eqn:reflectionTermEq}
   L_{amb} = \frac{k}{\pi} E_{amb},
\end{equation}
given the display's reflectivity $k$ (for common displays, $k < 0.01$) and the ambient illumination level $E_{amb}$ in units of lux. The final observed luminance is then equal to $L + L_{amb}$. With this extended model, we include the effect of varying ambient conditions on the viewing experience \cite{pdp, lumq}.

\subsection{Extending IQA to HDR}

% Perceptual uniformity implies that the human visual perception of color is consistent across the full color space. 
% A change of 10 steps in the sRGB color space appears approximately as noticeable to an observer across the full range of sRGB values, implying perceptual uniformity. 
% While most SDR metrics are calibrated to perceptually uniform pixel values, 

% trichromatic color values stored in HDR images are linearly related to luminance which humans perceive on a non-linear logarithmic scale.

% Humans perceive luminance on a logarithmic scale; trichromatic color values stored in HDR images are linearly related to luminance and thus are not perceptually uniform. By extension, 

% Most existing SDR metrics are calibrated to perceptually uniform pixel values, making them unsuitable for accurately processing perceptually non-uniform HDR values. 

\begin{figure}[t]
\centering
\includegraphics[width=\linewidth, trim={0.cm 0.2cm 0.cm 0.cm}]{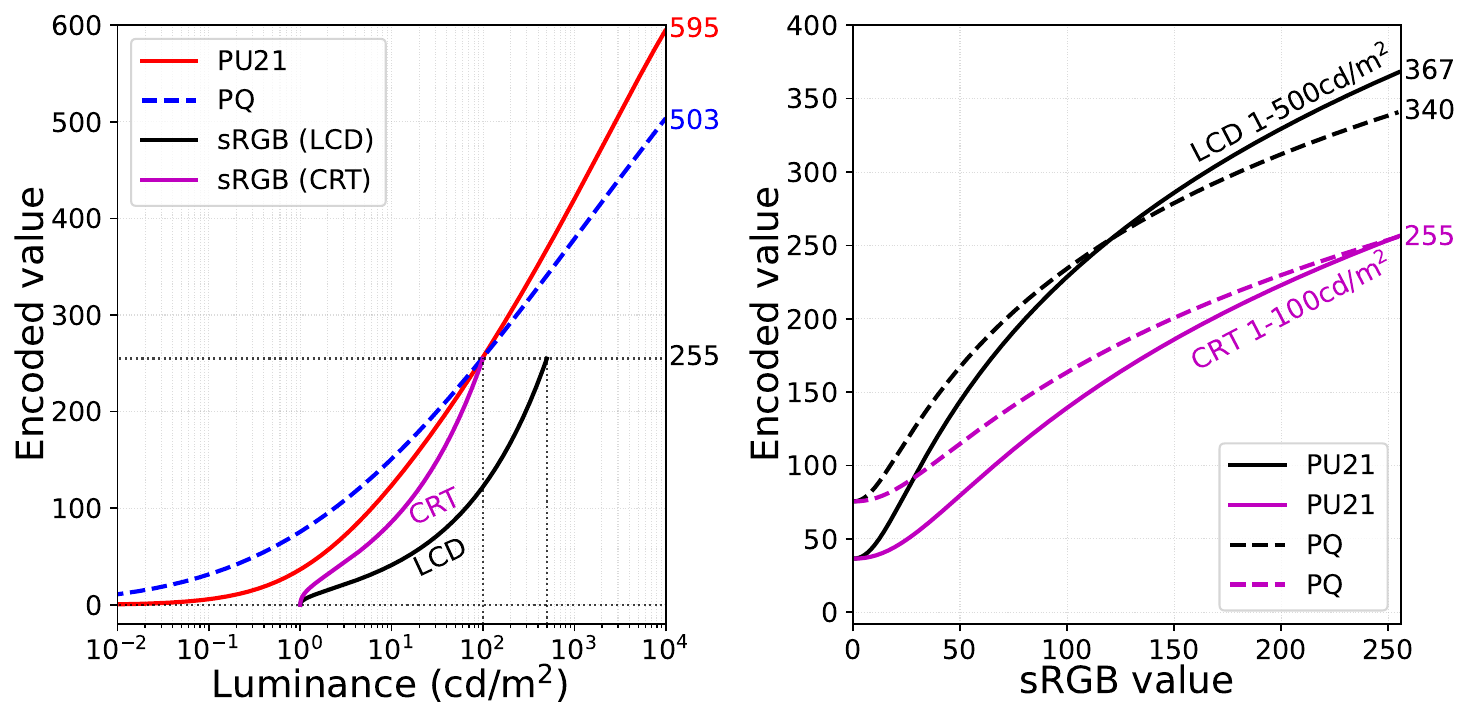}
\caption{Perceptually uniform encoding with PU21 \cite{pu21} (banding + glare variant) and PQ \cite{pq} (scaled by 255) contrasted with the approximate mapping between luminance and the sRGB non-linearity for typical CRT and LCD displays (simulated with \autoref{eqn:displayModel}). Left: the full range of encoded luminance. Right: the mapping between sRGB values and PU units. 
}
\label{fig:pu21_range}
\vspace{-15pt}
\end{figure}

Most existing SDR metrics are calibrated to perceptually uniform pixel values. On the other hand, trichromatic color values stored in HDR images are not perceptually uniform because they are linearly related to luminance, which humans perceive on a logarithmic scale. By extension, most SDR metrics cannot reliably predict image quality for HDR inputs. To improve the accuracy of SDR metrics on HDR data, perceptually uniform (PU) encoding \cite{puEncoding2008} first transforms luminance into approximately PU values by matching contrast detection thresholds across a wide range of luminance conditions. Existing SDR quality metrics, such as PSNR and SSIM, were shown to produce significantly more accurate predictions for PU-encoded HDR data. As illustrated in \autoref{fig:pu21_range}, PU21 encoding \cite{pu21} transforms luminance inputs of 0.005--10000 \candela to PU units. By design, luminance levels of 0.1--100 \candela (typical SDR display luminance) map to approximately 256 steps in the PU space, ensuring that SDR metrics produce comparable results for PU-encoded SDR data. 

Lastly, PQ \cite{pq} transforms luminance to a relatively PU space using similar derivations as PU21. While PU21 is concerned with the application of SDR metrics to HDR content, PQ is optimized to reduce visible quantization artifacts in HDR image formats, offering a coding scheme more aligned with human perception.

% Some methods are specifically designed for IQA on HDR content. Most notably, HDR-VDP \cite{hdrvdp2_2011, hdrvdp3}, 

% extended dynamic range affects the perceived visibility of image quality distortions. 
% The perceived quality of HDR images 

 % in absolute colorimetric units

\subsection{Domain Adaptation}

In machine learning, when labeled data is scarce (as is the case for HDR IQA), the common solution is to use other available datasets for closely related tasks. Naturally, as such source data may differ from the desired target domain, trained models may have suboptimal performance on the target data due to the problem of domain shift. To address this limitation, various domain adaptation techniques aim to facilitate the transfer of knowledge from a source domain to a target domain, mitigating the degradation in performance caused by domain shift and improving generalization on the target data \cite{easyDa, domainAdaptationBengio, domainAdaptationSurvey}. 
In this work, we focus specifically on deep feature activation CORrelation ALignment (CORAL) \cite{CORAL}, a DA technique that aligns the statistical properties of source and target distributions. With CORAL, neural networks are trained with an additional loss term defined as the distance between the second-order statistics of the two involved data distributions:
\begin{equation}
\label{eqn:coral}
  % \mathscr{L} % uses `mathrsfs`
  \mathcal{L}_{CORAL} = \frac{1}{4d^2} \|C_S - C_T\|^2_F,
\end{equation}
where $C_S$ and $C_T$ are the covariance matrices of the source and the target $d$-dimensional feature activations, respectively, and \hbox{$\|\cdot\|^2_F$} is the Frobenius norm. Optimizing CORAL loss leads to increased statistical similarity between the source and target domains, which in turn allows trained models to learn domain-invariant but task-specific features, consistently improving generalization on the target domain. 

% effective yet simple domain adaptation technique with proven results.

% , where a pre-trained model is applied to a secondary task or data. 
% , which align second-order statistics of source and target distributions to reduce the effect of domain shift.
% SDR algorithms can be adapted to HDR 
% Most quality metrics operate with sRGB pixel values: this represents the standard viewing experience on SDR displays.
% Given an input image as gamma-corrected pixel values, the parameters of the display, and the ambient illumination level, we simulate the signal that reaches the observer’s eye as physical luminance in photometric units of \candela as per Equation \ref{eqn:displayDegradationModel}:

\section{Methodology}

% Training on PU Units:
% 0-255 PU are very similar to sRGB
% PU units normalization from 0-255 to 0-1 instead of 0-maxRange (approx 600) to 0-1.

We investigate various modifications to the training procedure to adapt deep IQMs pre-trained on SDR data to HDR. First, to ensure out-of-the-box performance on PU-encoded SDR data for networks pre-trained with sRGB images, we verify a more intuitive normalization scheme which aligns PU-encoded SDR values to the range of sRGB inputs. Second, to further improve generalization on the target HDR domain, we fine-tune networks on a mix of SDR and HDR data with optional domain adaptation. Lastly, we ensure that the trained models perform well on both SDR and HDR data, making them more practical for real-world applications of IQA.
% where display characteristics and ambient conditions affect perceived image quality. 

% Given some simplifying assumptions about the used display systems, we existing SDR data 

% MAE loss + Deep CORAL loss as described bellow
% yielding a non-linear transformation which 

\subsection{Training with PU-encoded Data}

Similar to prior work on IQA for HDR \cite{puEncoding2008, pu21, upiq, lumq}, we represent gamma-encoded SDR and linear HDR color values on a unified scale in photometric units of luminance by computing the display response with \autoref{eqn:displayModel}. Following the specifications of the sRGB color space, SDR displays are modelled with \mbox{$L_{max}=100$ \candela} and $L_{blk}=0.5$ \candela (effective contrast ratio of 200:1). For HDR image formats, the stored trichromatic color values either directly encode luminance or can be tone-mapped to a given display characteristic using a similar approach. Physical luminance values are then encoded with PU21 \cite{pu21} for perceptual linearity and used to train neural networks.

To mitigate precision errors and stabilize training, neural networks are trained with values rescaled to a consistent floating-point range (e.g., 0.0--1.0 or similar).
% More formally, for a given input signal $x$, normalization is computed as follows:
% \begin{equation}
% \label{eqn:normalization}
%    x_{\text{normalized}} = (x - x_\text{min}) / (x_\text{max} - x_\text{min}) ,
% \end{equation}
% where $x_\text{min}$ and $x_\text{max}$ describe the original full range of values $x$.
PU-encoded values likewise need to be rescaled to some known range when used as input to a neural network. 
However, the choice of normalization scheme for PU-encoded values is not as intuitive.
% the full range of PU-encoded values to 0--1 is not intuitive for networks pre-trained on sRGB data. 
As illustrated in \autoref{fig:pu21_range}, PU21 (\textit{banding + glare} variant) maps luminance levels of 0.005--10000 \candela to roughly 0--595 PU units, with SDR luminance levels of 100 \candela explicitly scaled to 256 PU steps to ensure compatibility with conventional SDR metrics that operate on sRGB. To adhere to the original design of PU encoding, instead of normalizing by the full range of PU-encoded values as was done for training PU-PieAPP \cite{upiq}, we align 255 PU units (100 \candela) with 1.0, while PU-encoded values outside this range exceed 1.0 after normalization. We refer to this normalization scheme as ``\textit{255}'', because PU values are divided by 255 for normalization instead of the maximum PU encoded value $P_{max}$. SDR luminance levels of 0--100 \candela then map to 0--1, while HDR luminance levels of 100-10000 \candela) reach roughly 2.3 after normalization.
% $595/255\approx2.3$). 
% when normalizing PU-encoded values, we use $x_{max}=255$ instead of
% $f_{PU}(L_{m})$, where $f_{PU}$ is the PU21 transform and $L_{m}=10000$ \candela, i.e., the highest supported luminance value.

The choice of normalization scheme (divide by $P_{max}$ or \textit{255}) has an effect on initial and final performance levels. The benefit of \mbox{$P_{max}$} normalization is that PU-encoded values (SDR and HDR) match the range of input values used in pre-training; the downside---PU-encoded SDR inputs (originally as 0--1 sRGB values) are arbitrarily compressed to roughly 0--0.4 ($255/595\approx0.4$). With \textit{255} normalization, PU-encoded SDR luminance levels are aligned to sRGB range and networks pre-trained on sRGB data consequently produce similar predictions for PU-encoded SDR inputs. On the other hand, PU-encoded HDR luminance levels then map to a range of inputs unseen in training, emphasizing the domain expansion from SDR to HDR. Although a pre-trained network is not expected to produce reliable predictions on PU-encoded HDR data, it has guaranteed performance on SDR. 
We can then adapt pre-trained networks to the full range of PU-encoded data with additional fine-tuning and domain adaptation, which we hypothesize to be more effective under \textit{255} normalization.

\subsection{Domain Adaptation for HDR}

We incorporate domain adaptation in the training procedure to facilitate the transfer of knowledge between the SDR and HDR domains. To this end, we optimize deep CORAL loss \cite{CORAL} to align the correlations of deep-layer activations between SDR and HDR data. Under such a training regime, at each training iteration, we acquire a batch of SDR data and a batch of HDR data, compute network predictions and feature representations for the inputs, calculate conventional loss functions, e.g., mean average error (MAE), between the expected and predicted values, and finally the deep CORAL loss $L_{CORAL}$ between the SDR and HDR feature vectors. The loss terms are then summed as follows:
\begin{equation}
\label{eqn:lossWithCORAL}
    % \mathcal{L} = \mathcal{L}_{CLASS} + \lambda \mathcal{L}_{CORAL},
  \mathcal{L} = \alpha \mathcal{L}_{SDR} + \beta \mathcal{L}_{HDR} + \lambda \mathcal{L}_{CORAL},
\end{equation}
where $L_{SDR}$ and $L_{HDR}$ are conventional loss functions for IQA (we use MAE) on the SDR and the HDR batches, respectively, and $\alpha$, $\beta$, $\lambda$ are customizable weight parameters that allow for different trade-offs between the three defined losses. 
% Typically, DA is applied between labeled source and unlabeled target data. 
Since CORAL loss specifically adapts feature representations and does not require quality labels (equivalently, $\beta=0$ or $L_{HDR}=0$), we can leverage generic unlabeled HDR data which is more widely available than HDR IQA data. The models are then trained with DA to produce statistically similar features for SDR and HDR, but are not explicitly trained for IQA on HDR data. Lastly, the magnitude of $\mathcal{L}_{CORAL}$ varies according to the domains and the feature dimension $d$, hence $\lambda$ must be adjusted according to the use case. As in prior work \cite{CORAL}, we roughly match the magnitude of $L_{CORAL}$ with other loss terms at the end of the training. 
% When labeled data is available for HDR, it is possible to bias the models to HDR by setting $\alpha < \beta$, which may be relevant for unbalanced datasets with more SDR data. 
% We find that setting $\lambda=100$ yields good results in our case.

\section{Experiments and Results}

To evaluate the effect of our proposed training recipe, we re-train several deep neural network-based IQA models with and without our modifications. In what follows, we describe our training procedure and our performance evaluations, demonstrating the effectiveness of transfer learning and domain adaptation in the context of IQA on HDR images. 

% , effectively leading to a transfer of knowledge between the two domains. 
% CORAL loss assesses the statistical difference between the SDR and HDR domains. 
% a batch of SDR data and a batch of HDR from the available data
% we sample a batch of SDR data and a batch of HDR data

\subsection{Datasets} 

We test models on the UPIQ dataset \cite{upiq}, which consolidates two popular SDR IQA datasets (LIVE \cite{LIVEdatabase} and TID2013 \cite{tid2013}) and two smaller HDR IQA datasets (Korshunov \cite{korshunov} and Narwaria \cite{narwaria}), with quality labels realigned to a common scale (in JOD units) through additional subjective experiments. We train our metrics on IQA data from the larger KADID-10k dataset \cite{kadid10k} (SDR). 
For domain adaptation, we also use HDR images from an HDR image reconstruction dataset (not IQA), SI-HDR \cite{si-hdr}. 
We summarize relevant details on the used datasets in \autoref{table:datasets}.

\begin{table}[t]
\begin{center}  
\caption{Comparison of the used datasets.}
\label{table:datasets}
\vspace{10pt}  % too much between table and title
\footnotesize
\setlength\tabcolsep{3pt}
\begin{tabular}{c|ccccc}
\Xhline{2\arrayrulewidth}
Dataset & No. Ref.  & No.   & No. Dist. & Resolution    & Dynamic \\
Name    & images    & Dist. & images    & (h$\times$w)  & range \\
\hline
LIVE \cite{LIVEdatabase} & 29 & 5 & 779 & 512$\times$768 & SDR\\
TID2013 \cite{tid2013} & 25 & 24 & 3,000 & 384$\times$512 & SDR\\ 
KADID-10k \cite{kadid10k} & 81 & 25 & 10125 & 384$\times$512 & SDR\\
Narwaria \cite{narwaria} & 25 & 2 & 140 & 1080$\times$1920 & HDR\\ 
Korshunov \cite{korshunov} & 25 & 3 & 240 & 1080$\times$944 & HDR\\ 
\hline
SI-HDR \cite{si-hdr} & 181 & N/A & N/A & 1080$\times$1888 & HDR\\
\Xhline{2\arrayrulewidth}
\end{tabular}
\vspace{-20pt}  % for some reason there is 10pt of extra space below the tables
% \vspace{-10pt}  % reduce space 
\end{center} 
\end{table}

\subsection{Experimental Setup}

\begin{figure}[t]
\centering
\includegraphics[width=\linewidth, trim={0.cm 0.35cm 0.cm 0.cm}]{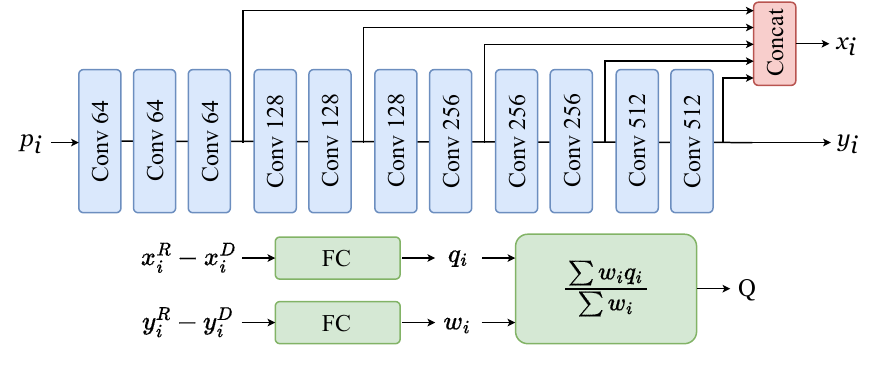}
\caption{Diagram of PieAPP quality metric \cite{pieapp}. For each $64 \times 64$ patch, deep feature activations from 5 convolutional layers are computed and concatenated; a fully-connected layer predicts the quality score given the difference between the reference and the distorted patch features. Features from the last layer are used to predict patch weights. The final image quality is computed as a weighted sum of patch-wise scores. Adapted from \cite{pieapp}.}
\label{fig:pieapp}
\vspace{-10pt}
\end{figure}

\begin{figure}[t]
\centering
\includegraphics[width=0.9\linewidth, trim={0.cm 0.2cm 0.cm 0.cm}]{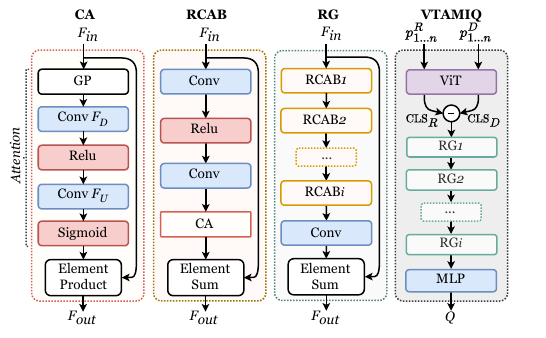}
\caption{Diagram of VTAMIQ \cite{vtamiq}. Patches from the reference and the distorted images are encoded by Vision Transformer (ViT) \cite{vit2020}, the corresponding CLS token difference is computed and calibrated by a series of residual groups (RGs) based on channel attention (CA) modules. A fully-connected layer (MLP) predicts the final quality score. Adapted from \cite{vtamiq}.}
\label{fig:vtamiq}
\vspace{-10pt}
\end{figure}

% \subsubsubsection{Tested IQA models}
In previous work on IQA for HDR \cite{upiq}, the CNN-based PieAPP quality metric \cite{pieapp} (model architecture depicted in \autoref{fig:pieapp}) was trained from scratch on PU-encoded SDR and HDR data normalized to range 0--1 ($P_{max}$), with the HDR variant termed PU-PieAPP. 
We re-train PU-PieAPP using our training recipe
consisting of pre-training on SDR data and fine-tuning on PU-encoded data with optional domain adaptation, following the optimization criterion defined in \autoref{eqn:lossWithCORAL}.
Under similar setup, we also re-train VTAMIQ \cite{vtamiq}, a transformer-based FR IQA model extended for multi-scale patch processing via scale embedding \cite{musiq} (architecture in \autoref{fig:vtamiq}), with the final model analogously termed PU-VTAMIQ. 
% We train models for IQA in several stages consisting of pre-training and fine-tuning. 
For both metrics, we use feature extraction networks pre-trained for classification of sRGB images. For PieAPP, we initialize the feature extractor with pre-trained weights for VGG16 \cite{veryDeepCnns}\footnote{PieAPP's CNN feature extractor architecture is similar to VGG16}. For VTAMIQ, we use a pre-trained vision transformer. 

% We train the metrics for IQA on sRGB data from the conventional SDR datasets. We then fine-tune networks on PU-encoded data using a mix of SDR and HDR samples, optionally with domain adaptation. Under such setting, we optimize the loss functions defined in \autoref{eqn:lossWithCORAL}. 

Both PieAPP and VTAMIQ use patch inputs, which we obtain by tiling input images with randomized perturbations\footnote{Sampling details in \url{https://github.com/ch-andrei/VTAMIQ}}, resulting in a relatively uniform yet randomized view of the full image.
% an image is split onto a grid of cells and patches are randomly sampled from each cell leading to more consistent coverage. 
For PieAPP, we sample 128 patches of size $64\times64$ for each image during training and 1024 patches for testing. For VTAMIQ, we use a variant of ViT extended with scale embedding \cite{musiq}: patches are sampled at five different scales with initial patch sizes $p \in \{16, 32, 64, 128, 256\}$ pixels (downsampled to $16\times16$ when input to ViT), and with 512 patches for training and 2048 for testing. 

For domain adaptation, we apply deep CORAL to the feature vectors produced by the tested models for SDR and HDR data. While PieAPP computes CNN features and the corresponding quality scores for each patch independently, VTAMIQ produces a single compact feature representation and quality score by jointly encoding all patches with the CLS token. For PieAPP, we compute CORAL loss on the features from the last Conv512 layer ($y_i$ in \autoref{fig:pieapp}) with flattened feature dimension $d=2048$. We concatenate the feature vectors for all patches for the reference and the distorted images (latter, if available). As an example, for a batch size of 8 image pairs with 64 patches per image, we will compute the CORAL loss between two feature matrices of size $1024\times2048$ ($8\times64\times2=1024$ and $d=2048$). Similarly for VTAMIQ, we concatenate the CLS tokens with $d=768$ for all reference and distorted images in a batch; for batch size and patch count above, CORAL loss is computed between $16\times768$ feature matrices.

% \subsubsubsection{Implementation Details} 
Our implementation is in Pytorch \cite{pytorch}, with all training performed on a single NVIDIA GeForce RTX 3090 GPU with 24GB of video memory. We use the AdamW optimizer \cite{adamw} with the recommended parameters and an initial learning rate of $10^{-4}$, exponentially decayed at the end of each epoch with a final goal of $10^{-6}$. Practical training times depend on the used dataset, but we generally train for 50 epochs for each run. Since we leverage pre-training, our models converge faster than PU-PieAPP from \cite{upiq} (trained for 500 epochs). Note that while the original PU-PieAPP \cite{upiq} was trained on data encoded with \mbox{PU08 \cite{puEncoding2008}}, we use the updated PU21 \cite{pu21}, specifically its \textit{banding + glare} variant. 
% We pre-train for 25 epochs on KADID-10k and train for 25 epochs on each tested subset of UPIQ. 

\subsection{Performance Evaluations}

\newcommand{\pie}{PieAPP\xspace}
\newcommand{\pupie}{\mbox{PU-PieAPP}\xspace}
\newcommand{\pupieupiq}{\mbox{PU-PieAPP*}\xspace}
\newcommand{\puvit}{\mbox{PU-VTAMIQ}\xspace}

We evaluate the tested models on the UPIQ dataset with 5-fold cross-validation, splitting the available data into train, validation, and testing sets with a splitting ratio of 60-20-20 across the reference image dimension. When training on UPIQ, we take extra care to correctly isolate reference images from LIVE and TID2013 subsets of UPIQ (overlapping images are assigned to TID2013) and split the SDR and HDR portions in roughly the same ratio. 
% We also evaluate cross-dataset performance, by testing on the different subsets of UPIQ; when training (if applicable), the given test subset is held out. 
Following common practice, we assess the performance of the tested models with Spearman rank order correlation coefficient (SROCC) and Pearson linear correlation coefficient (PLCC) between the expected and the predicted quality scores. As in prior work, a logistic fit is applied before computing PLCC \cite{sheikhIqaFit}.

\subsubsubsection{Pre-training and fine-tuning} 
% \subsection{Fine-tuning and Normalization}
% \subsubsubsection{Pre-training on sRGB data} 
We test metrics pre-trained on sRGB data from \mbox{KADID-10k} directly on PU-encoded data from UPIQ. 
% We also verify the effect of different normalization schemes (NS).
As presented in \autoref{tab:results}, without fine-tuning on PU-encoded data, PieAPP and VTAMIQ produce accurate predictions for PU-encoded SDR data (0.87--0.91 SROCC) but are only as reliable on HDR data as the PU variants of conventional IQMs (0.60--0.76 SROCC).
Conversely, HDR-VDP \cite{hdrvdp2_2011} and HDR-VQM \cite{hdrvqm}, offer very strong performance on the HDR subsets, but are suboptimal for SDR. 
We then fine-tune VTAMIQ and \pie on PU-encoded data from KADID-10k to produce their respective PU variants, \puvit and \pupie. We note the resulting improvement on both SDR and HDR content. Even only training on PU-encoded SDR data improves performance on the HDR subset. VTAMIQ-based models consistently outperform PieAPP, nearly matching HDR-VDP on the HDR subset (0.81 SROCC).

We contrast the performance of our models with \pupieupiq trained in \cite{upiq} from scratch on data from UPIQ encoded with PU08 \cite{puEncoding2008} and under $P_{max}$ normalization. \pupieupiq has stronger cross-validation performance on the \textit{Full} set of UPIQ, which is expected, because the model is directly trained on subsets of UPIQ, while our models are trained on KADID-10k. On the other hand, when \pupieupiq is trained on the HDR subset of UPIQ and evaluated on its SDR subset, it performs poorly, though its performance on the HDR subset, when trained on SDR, is only slightly lower than ours. Our models thus provide a more optimal balance of performance on both subsets, demonstrating the benefit of transfer learning, where adequate pre-training improves generalization on unseen data for both SDR and HDR inputs. We achieve further performance improvements on HDR inputs with additional fine-tuning and domain adaptation.

\newcommand{\cmk}{\checkmark}
\newcommand{\B}[1]{\textbf{#1}}
\newcommand{\U}[1]{\underline{#1}}

\begin{table}[!t]
\centering
\caption{Performance (SROCC and PLCC) on subsets of \mbox{UPIQ}. {\normalfont \Large  For \textit{Full} set, we report 5-fold cross-validation performance; for \textit{SDR} and \textit{HDR} subsets, we test on the entire subset. We train PieAPP and VTAMIQ on sRGB data, with PU-PieAPP and PU-VTAMIQ further fine-tuned on PU-encoded SDR data. 
Other metrics are as reported in \cite{upiq}, where PU-PieAPP* is trained on UPIQ (with test set held out). Column \textit{Input} describes input type and normalization scheme. Best scores \textbf{bolded}, second best \underline{underlined}.
% PU-PieAPP and PU-VTAMIQ are further fine-tuned on PU data with domain adaptation. 
}}
\label{tab:results}
\vspace{12pt}
\footnotesize
\setlength\extrarowheight{0.5pt}
\setlength\tabcolsep{4pt}
\begin{tabular}{l|l|cc|cc|cc}
\Xhline{2\arrayrulewidth}
\multirow{2}{*}{\textbf{Method}} & \multirow{2}{*}{\textbf{Input}} & \multicolumn{6}{c}{\textbf{Tested subset of UPIQ}} \\
\cline{3-8} 
 &&
  \multicolumn{2}{c|}{\textbf{Full}} &
  \multicolumn{2}{c|}{\textbf{SDR}} &
  \multicolumn{2}{c}{\textbf{HDR}} \\  
\Xhline{2\arrayrulewidth}
% \hline
% \hline
FSIM        & Luminance    & 0.82 & 0.89 & 0.54 & 0.51 & 0.45 & 0.34 \\
PU-FSIM     & PU08        & 0.84 & 0.90 & 0.77 & 0.77 & 0.71 & 0.66 \\
HDR-VDP     & Luminance    & 0.82 & 0.84 & 0.82 & 0.78 & \U{0.81} & 0.72 \\
HDR-VQM     & Luminance    & 0.78 & 0.82 & 0.60 & 0.62 & \B{0.87} & \B{0.86} \\
% \hline
% PU-PieAPP*  & PU08 ($P_{max}$)        & 0.94 & 0.96 & 0.65 & 0.67 & 0.74 & 0.73 \\
\hline
\multicolumn{8}{c}{\textit{Trained on KADID-10k (sRGB)}} \\
\hline
\multirow{2}{*}{PieAPP} 
            & PU21 ($P_{max}$)     & 0.85 & 0.84 & 0.87 & 0.87 & 0.63 & 0.65 \\
            & PU21 (\textit{255})   & 0.86 & 0.85 & 0.88 & 0.89 & 0.60 & 0.63 \\
\hline
\multirow{2}{*}{VTAMIQ} 
            & PU21 ($P_{max}$)     & 0.87 & 0.87 & 0.90 & 0.91 & 0.76 & 0.76 \\
            & PU21 (\textit{255})   & 0.88 & 0.89 & 0.91 & 0.92 & 0.71 & 0.72 \\
\hline
\multicolumn{8}{c}{\textit{Fine-tuned on KADID-10k (PU-encoded)}} \\
\hline
\multirow{2}{*}{PU-PieAPP} 
            & PU21 ($P_{max}$)     & 0.87 & 0.87 & 0.89 & 0.89 & 0.75 & 0.75 \\
            & PU21 (\textit{255})   & 0.87 & 0.86 & 0.90 & 0.90 & 0.63 & 0.66 \\
\hline
\multirow{2}{*}{PU-VTAMIQ} 
            & PU21 ($P_{max}$)     & 0.89 & 0.90 & \U{0.91} & \U{0.92} & \U{0.81} & \U{0.81} \\
            & PU21 (\textit{255})   & \U{0.90} & \U{0.91} & \B{0.93} & \B{0.94} & 0.72 & 0.73 \\
\hline
\multicolumn{8}{c}{\textit{Trained on UPIQ (PU-encoded)}} \\
\hline
PU-PieAPP*  & PU08 ($P_{max}$)  & \B{0.94} & \B{0.96} & 0.65 & 0.67 & 0.74 & 0.73 \\
\Xhline{2\arrayrulewidth}
\end{tabular}
\vspace{-15pt}
\end{table}

% We aim to produce a different trade-off for SDR-HDR performance. To this end, we 
% by fine-tuning PieAPP and VTAMIQ on PU-encoded data 
% with optional domain adaptation between SDR and HDR. 

\subsubsubsection{Normalization scheme for PU-encoded values} We find that the performance of the trained metrics increases on SDR and decreases on HDR data when \textit{255} normalization is used instead of $P_{max}$. This is expected because PU-encoded SDR luminance levels appear more similar to sRGB values when normalizing by \textit{255}, while HDR signals exceed the range of data used in pre-training. However, we confirm that $P_{max}$ normalization results in significantly better final performance, discouraging the use of \textit{255} normalization, despite our original motivation of leveraging the similarity between PU-encoded SDR and sRGB. Since our main objective is HDR performance, where \textit{255} normalization is suboptimal, we emphasize our results under $P_{max}$ normalization. 

% While \textit{255} normalization allows for better performance on SDR data, our main objective is HDR, hence $P_{max}$ does not suit our needs.

\subsubsubsection{Training with domain adaptation} 
We consider three domain adaptation configurations with different training procedures and source-target domains. While we use labeled SDR IQA data from KADID-10k (PU-encoded) as the source domain, the HDR-like target domain can be:
(i) unlabeled authentic HDR images (\mbox{$S\xrightarrow{}H_U$}),
(ii) labeled synthetic HDR-like images simulated from sRGB images in SDR IQA datasets  (\mbox{$S\xrightarrow{}H_S$}), and
(iii) labeled authentic HDR images from HDR IQA datasets (\mbox{$S\xrightarrow{}H_L$}).
% \begin{enumerate}[label=(\roman*), nosep]
%     % \setlength\itemsep{-5pt}
%     \item unlabeled HDR images (\mbox{$S\xrightarrow{}H_U$}),
%     \item labeled synthetic HDR-like images simulated from sRGB images in SDR IQA datasets  (\mbox{$S\xrightarrow{}H_S$}),
%     \item labeled authentic HDR images from HDR IQA datasets (\mbox{$S\xrightarrow{}H_L$}).
% \end{enumerate} 
Option \mbox{$S\xrightarrow{}H_U$} follows the common setting for DA with unlabeled target data, produces subtle improvements in generalization on the target domain, but typically does not outperform training on labeled data.
With option \mbox{$S\xrightarrow{}H_S$}, we provide additional optimization guidance with labeled HDR-like data simulated from the more abundant SDR IQA data, granted the distribution of luminance values does not truly come from an HDR source and the reused SDR quality labels are perhaps not as reliable for HDR.
Lastly, with \mbox{$S\xrightarrow{}H_L$}, although authentic HDR IQA data is used, its severely limited availability poses practical challenges due to overfitting and noisy evaluations. 

For $S\xrightarrow{}H_U$, we experiment with (i) HDR images from the SI-HDR dataset, (ii) SDR images from KADID-10k simulated as HDR, and (iii) SDR images from UPIQ simulated as HDR (UPIQS). 
For $S\xrightarrow{}H_S$, we train on labeled data from \mbox{KADID-10k} simulated as HDR.
% For \mbox{$S\xrightarrow{}H_U$}, we use unlabeled HDR images from the SI-HDR dataset or the HDR images from UPIQ (minus the quality labels). 
To generate synthetic HDR-like IQA data, we reuse the labels and images from SDR IQA datasets (e.g., KADID-10k or the SDR subset of UPIQ), but simulate HDR-like display response by controlling the $L_{max}$ parameter in \autoref{eqn:displayModel}. We sample $L_{max}$ from a normal distribution $\mathcal{N}(100, 10)$ for SDR and $\mathcal{N}(5000, 500)$ for HDR display response for additional data augmentation, instead of using a constant value as was done in \cite{upiq}. We apply a similar method to tone-map the linear HDR color values from SI-HDR. 
Lastly, for $S\xrightarrow{}H_L$, we train with cross-validation on labeled HDR data from UPIQ (UPIQH) and test on a held out set, which makes comparison with other DA configurations problematic, but nevertheless showcases the benefit of CORAL.

\begin{table}[t]
% \vspace{-10pt}  % too much space above table
\begin{center}  
\caption{Performance (SROCC) on the HDR subset of UPIQ for different training and domain adaptation configurations.
{\normalfont \Large 
% Column \textit{Input} describes normalization scheme. 
All runs use PU21 encoding and $P_{max}$ normalization.
Training with CORAL loss indicated in column $\lambda$. For $S\xrightarrow{}H_U$, training with no CORAL loss is equivalent to results in \autoref{tab:results}, because only the SDR labels are used. Unlike other DA configurations, for $S\xrightarrow{}H_L$, we apply 5-fold cross-validation on the HDR subset of UPIQ.
% Performance reported $S\xrightarrow{}H_L$
% Performance for PU-PieAPP* as reported in \cite{upiq}.
}
% PU-PieAPP and PU-VTAMIQ are further fine-tuned on PU-encoded data (UPIQ) with the test subset held out.
% }
}
\label{tab:results_da_coral}
\vspace{12pt}  % not enough space below title
\footnotesize
\setlength\tabcolsep{3.5pt}
\setlength\extrarowheight{0.5pt}
\begin{tabular}{l|c|ccc|c||c}
\Xhline{2\arrayrulewidth}
\multirow{3}{*}{\textbf{Method}} & 
% \multirow{3}{*}{\textbf{Input}} & 
\multirow{3}{*}{\textbf{$\lambda$}}& \multicolumn{5}{c}{\textbf{DA Configuration and Target}} \\ \cline{3-7} 
& & \multicolumn{3}{c|}{\textbf{$S\xrightarrow{}H_U$}} & \textbf{$S\xrightarrow{}H_S$} & \textbf{$S\xrightarrow{}H_L$} \\ \cline{3-7} 
&       & SIHDR & KADID & UPIQS & KADID & UPIQH \\
\hline 
% \hline 
\multirow{2}{*}{PU-PieAPP}   
&       & 0.75 & 0.75 & 0.75 & 0.78 & 0.85 \\
& \cmk  & 0.76 & 0.78 & 0.76 & 0.79 & 0.86 \\
\hline
\multirow{2}{*}{PU-VTAMIQ} 
&       & 0.81 & 0.81 & 0.81 & 0.88 & 0.89 \\
& \cmk  & 0.82 & 0.85 & 0.84 & 0.89 & 0.91 \\
% \hline
% \multirow{2}{*}{PU-PieAPP}   
% &\textit{255}&       & 0.63 & 0.63 & 0.63 & ???? & ???? \\
% &\textit{255}& \cmk  & ???? & ???? & ???? & ???? & ???? \\
% \hline
% \multirow{2}{*}{PU-VTAMIQ} 
% &\textit{255}&      & 0.72 & 0.72 & 0.72 & ???? & ???? \\
% &\textit{255}& \cmk  & ???? & ???? & ???? & ???? & ???? \\
\Xhline{2\arrayrulewidth}
\end{tabular}
\vspace{-25pt}  % for some reason there is 10pt of extra space below the tables
\end{center} 
\end{table}

Our performance evaluations on the HDR subset of UPIQ for models trained with DA are presented in \autoref{tab:results_da_coral}, which we contrast with \autoref{tab:results}, where the training is done without DA. We determine that training with CORAL yields subtle but noticeable performance improvements on the target HDR data for all tested DA configurations with both unlabeled and labeled target data. For $S\xrightarrow{}H_U$, training without CORAL loss is equivalent to training on KADID-10k (see \autoref{tab:results}). 
Our best performance is achieved with option $S\xrightarrow{}H_S$ with for KADID-10k as target, where we simulate synthetic HDR-like data. PU-VTAMIQ trained with $S\xrightarrow{}H_S$ has 0.89 SROCC on the HDR subset of UPIQ, outperforming HDR-VQM which has 0.87 SROCC; PU-PieAPP, generally, reacts less favorably to DA with CORAL. To isolate the effect of optimizing deep CORAL loss, we also train without the CORAL loss term, i.e., only using the labeled SDR and HDR data but without DA between them. While the main improvement comes from having labeled HDR-like training data, CORAL loss leads to additional albeit subtle gains. Finally, while we experiment with DA using $P_{max}$ and \textit{255} normalization schemes, we find that $P_{max}$ offers stronger performance for all setups, unless we train directly on HDR data from UPIQ ($S\xrightarrow{}H_L$), where the final performance is comparable. For brevity, we omit our DA results for \textit{255} normalization.

\section{Discussion}

% Domain adaptation between SDR and HDR is an underexplored topic, especially for IQA. 
% Although our paper focuses primarily on IQA, we expect similar performance improvements if our training strategy is applied to other HDR tasks. 
% With our modifications to the PU-encoded inputs, namely the scaling of SDR and HDR data, we ensure compatibility on SDR data, but also emphasize the domain change for HDR values.
% We perform domain adaptation on the deep feature activations and train models to produce similar statistical properties for SDR and HDR data; CORAL loss specifically targets second order statistics of the feature maps. With this, knowledge on the SDR domain is transferred to HDR applications.

With transfer learning, a model is trained on one task and repurposed for a different but related task. Domain adaptation complements pre-training by facilitating transfer of knowledge from source to target domain. We train networks on sRGB data and adapt them to PU-encoded HDR data with additional fine-tuning and domain adaptation between SDR and HDR. 
Fine-tuning on PU-encoded data produces a considerable improvement to generalization on HDR content.
DA contributes to an additional incremental gain in IQA performance on the target HDR domain. 
% Domain adaptation between SDR and HDR is an underexplored topic. 
We explored DA with unlabeled and labeled target data, achieving higher performance in both scenarios.
% While we focus on IQA, we expect our findings to also apply to other HDR tasks, where limited labeled task-specific data is available.
% \subsubsubsection{Importance of IQA data}
For unlabeled DA for IQA ($S\xrightarrow{}H_U$), we note the importance of distorted images in the target dataset. Although \mbox{SI-HDR} contains authentic HDR images, it produce meager improvement when used as DA target, which we hypothesize is due to its relatively limited size and lack of distorted images. Conversely, SDR images from KADID-10k contain image quality distortions and more variation, resulting in improved performance when used as DA target. Granted, \mbox{SI-HDR} only contains 181 images, which may be insufficient for our application---we leave DA on a larger unlabeled HDR dataset to future work. While we focus on IQA, we expect our findings to also apply to other HDR tasks, where limited labeled task-specific data is available.

% The task of IQA requires knowledge of the distortions present in a given distorted image; while the magnitude of perceived distortions may change between SDR and HDR, the overall IQA process remains similar.

We note, however, that DA adds complexity to the training procedure and presents certain challenges. First, with deep CORAL, the weight of the CORAL loss must be tuned to a given use-case: despite recommendations in the original work, this potentially requires extensive empirical experimentation. Secondly, CORAL loss assumes that the second-order statistics of two data distributions can be aligned without limiting domain-specific representation, instead leading to task-specific but domain-invariant learning, but there is no guarantee that this holds for all tasks and data distributions. Moreover, while CORAL of deep feature representations is originally intended for other vision tasks, in IQA, the distortions arguably matter as much as natural image statistics. Using unlabeled and undistorted images for DA is perhaps not enough to transfer knowledge of relevant features for IQA. In that regard, it is then difficult to apply DA in its original form, hence why we find that CORAL is useful as an additional optimization criterion along regular IQA losses on both the source and target domains. Lastly, training with DA requires certain implementation details and data schedules to be changed in order to accommodate DA. This makes comparison against previous work problematic, as potential performance difference may be due to different training procedures. While we addressed this by training specifically with and without CORAL loss for similar data splits and other loss functions, there may be other details that affected the final performance. 

% training using exactly the same data splits, data sampling between SDR and HDR subsets, and other implementation-specific details 
% there needs to be a relative balance between the samples from the two domains. 

% that it assumes that the second-order statistics of two data distributions can be aligned without limiting domain-specific representation, instead leading to task-specific but domain-invariant learning. 

% Moreover, because DA relies on transferring knowledge from a source domain to a target domain, transferable knowledge may be limited if the performance on the source domain is inadequate

% While the HDR domain is our ultimate target, we also assume that a similar task, and consequently data, is more widely available in the SDR domain.
% rk does not adequate performance on the source domain.

\section{Conclusion}

We investigate more effective training strategies to adapt networks pre-trained on SDR data to HDR applications. First, we verify the effect of normalization schemes when training on data encoded with perceptually uniform transforms such as PU21. 
% PU-encoding by design aligns SDR data to the range of values in sRGB, improving the compatibility of models pre-trained on SDR data. 
When using networks pre-trained on sRGB data, we find that the benefit of aligning the full range of PU values to the range of values used in pre-training outweighs the benefit of similarity between PU-encoded SDR values and sRGB. Moreover, networks pre-trained on sRGB data produce excellent baselines for fine-tuning on PU-encoded data.
% , with adequate performance on HDR data.
% even PU-encoded SDR offers an additional performance boost.
With additional training and optional domain adaptation, we further consolidate IQA for SDR and HDR, leveraging the more widely available SDR data to transfer task-specific but domain-invariant knowledge from SDR (source domain) to HDR (target domain). Our results demonstrate that our combined training recipe offers much quicker convergence and stronger generalization on both SDR and HDR data.
% improved final prediction accuracy, and potentially different trade-offs for SDR-HDR performance.
% speeds up training and 
% leading to improved prediction accuracy and quicker convergence. 
Lastly, although we focus on IQA, our strategies likely extend to other tasks in HDR imaging. 

% \section{Acknowledgments}

% This work is a collaboration with Faurecia IRYStec Inc.

% Furthermore, to maximize the compatibility of networks pre-trained on SDR data, we modify the normalization procedure for data encoded with perceptually uniform transforms such as PU21 by aligning PU-encoded SDR data to the range of values in sRGB. 

% Together, our training strategies produce trained networks with more robust performance in the HDR domain. As physical displays, especially in HDR, are better described with linear photometric luminance -- 

%\section{Acknowledgments} 
%add the acknowledgement section here


%%%%%%%%%%%%%%%%%%%%%%%%%%%%%%%%%%
% Bibliography
%%%%%%%%%%%%%%%%%%%%%%%%%%%%%%%%%%

% \scriptsize
\small
\begin{spacing}{0.5}
\bibliographystyle{unsrt}
\bibliography{main}
\end{spacing}


% \begin{biography}
% Andrei Chubarau is a PhD candidate at McGill University. His research interests include computer vision and graphics.

% Tara Akhavan is the founder and CTO of IRYStec Software Inc. She holds a bachelor’s degree in computer engineering, a master’s degree in artificial intelligence, and a Ph.D. in image processing and computer vision from the Vienna University of Technology in Austria. Akhavan is vice chair of marketing for the Society of Information Display.

% Hyunjin Yoo is a senior research engineer and team lead at Faurecia IRYStec Inc. She received an M.S. in information and communications and a Ph.D. in information and mechatronics from Gwangju Institute of Science and Tech-nology in Gwangju, South Korea.

% James Clark received his BASc and PhD in Electrical Engineering from the University of British Columbia (1980,85). From 1985-1994 he was with the Division of Applied Sciences at Harvard University, and from 1994-96 was a visiting scientist at Nissan Cambridge Basic Research. Since then he has been with the department of Electrical and Computer Engineering at McGill University. His work has focused on computer vision and image processing. He is a senior member of IEEE.
% \end{biography}

\end{document}